\documentclass[10pt,twocolumn,letterpaper]{article}

\usepackage{cvpr}              
\usepackage{url}
\usepackage{booktabs}       
\usepackage{amsfonts}       
\usepackage{nicefrac}       
\usepackage{microtype}      
\usepackage{xcolor}         
\usepackage{caption}
\usepackage{graphicx}
\usepackage{amsmath}
\usepackage{amssymb}
\usepackage{booktabs}
\usepackage{verbatim}
\usepackage{color}
\usepackage{float}
\usepackage{epsfig}
\usepackage{caption,subcaption}
\usepackage{soul}
\usepackage{colortbl}
\usepackage[ruled, vlined]{algorithm2e}

\usepackage[symbol]{footmisc}
\usepackage{multirow}
\definecolor{cvprblue}{rgb}{0.21,0.49,0.74}
\usepackage[pagebackref,breaklinks,colorlinks,citecolor=cvprblue]{hyperref}

\def\paperID{1429} 
\def\confName{CVPR}
\def\confYear{2025}

\title{HOIGPT: Learning Long-Sequence Hand-Object Interaction with Language Models}

\author{Mingzhen Huang$^{1,2}$\footnote{Work done when interning at Meta}, Fu-Jen Chu$^1$, Bugra Tekin$^1$, Kevin J. Liang$^1$, Haoyu Ma$^1$, Weiyao Wang$^1$ \\ Xingyu Chen$^1$, Pierre Gleize$^1$, Hongfei Xue$^2$, Siwei Lyu$^2$, Kris Kitani$^1$, Matt Feiszli$^1$, Hao Tang$^1$
\\ 
$^1$ FAIR, Meta  $\quad  \quad ^2$ State University of New York at Buffalo 
}

\begin{document}
\maketitle
\def\mA{\mathcal{A}}
\def\mB{\mathcal{B}}
\def\mC{\mathcal{C}}
\def\mD{\mathcal{D}}
\def\mE{\mathcal{E}}
\def\mF{\mathcal{F}}
\def\mG{\mathcal{G}}
\def\mH{\mathcal{H}}
\def\mI{\mathcal{I}}
\def\mJ{\mathcal{J}}
\def\mK{\mathcal{K}}
\def\mL{\mathcal{L}}
\def\mM{\mathcal{M}}
\def\mN{\mathcal{N}}
\def\mO{\mathcal{O}}
\def\mP{\mathcal{P}}
\def\mQ{\mathcal{Q}}
\def\mR{\mathcal{R}}
\def\mS{\mathcal{S}}
\def\mT{\mathcal{T}}
\def\mU{\mathcal{U}}
\def\mV{\mathcal{V}}
\def\mW{\mathcal{W}}
\def\mX{\mathcal{X}}
\def\mY{\mathcal{Y}}
\def\mZ{\mathcal{Z}} 

\def\bbN{\mathbb{N}} 
\def\bbR{\mathbb{R}} 
\def\bbP{\mathbb{P}} 
\def\bbQ{\mathbb{Q}} 
\def\bbE{\mathbb{E}}

\def\1n{\mathbf{1}_n}
\def\0{\mathbf{0}}
\def\1{\mathbf{1}}

\def\A{{\bf A}}
\def\B{{\bf B}}
\def\C{{\bf C}}
\def\D{{\bf D}}
\def\E{{\bf E}}
\def\F{{\bf F}}
\def\G{{\bf G}}
\def\H{{\bf H}}
\def\I{{\bf I}}
\def\J{{\bf J}}
\def\K{{\bf K}}
\def\L{{\bf L}}
\def\M{{\bf M}}
\def\N{{\bf N}}
\def\O{{\bf O}}
\def\P{{\bf P}}
\def\Q{{\bf Q}}
\def\R{{\bf R}}
\def\S{{\bf S}}
\def\T{{\bf T}}
\def\U{{\bf U}}
\def\V{{\bf V}}
\def\W{{\bf W}}
\def\X{{\bf X}}
\def\Y{{\bf Y}}
\def\Z{{\bf Z}}

\def\a{{\bf a}}
\def\b{{\bf b}}
\def\c{{\bf c}}
\def\d{{\bf d}}
\def\e{{\bf e}}
\def\f{{\bf f}}
\def\g{{\bf g}}
\def\h{{\bf h}}
\def\i{{\bf i}}
\def\j{{\bf j}}
\def\k{{\bf k}}
\def\l{{\bf l}}
\def\m{{\bf m}}
\def\n{{\bf n}}
\def\o{{\bf o}}
\def\p{{\bf p}}
\def\q{{\bf q}}
\def\r{{\bf r}}
\def\s{{\bf s}}
\def\t{{\bf t}}
\def\u{{\bf u}}
\def\v{{\bf v}}
\def\w{{\bf w}}
\def\x{{\bf x}}
\def\y{{\bf y}}
\def\z{{\bf z}}

\def\balpha{\mbox{\boldmath{$\alpha$}}}
\def\bbeta{\mbox{\boldmath{$\beta$}}}
\def\bdelta{\mbox{\boldmath{$\delta$}}}
\def\bgamma{\mbox{\boldmath{$\gamma$}}}
\def\blambda{\mbox{\boldmath{$\lambda$}}}
\def\bsigma{\mbox{\boldmath{$\sigma$}}}
\def\btheta{\mbox{\boldmath{$\theta$}}}
\def\bomega{\mbox{\boldmath{$\omega$}}}
\def\bxi{\mbox{\boldmath{$\xi$}}}
\def\bnu{\mbox{\boldmath{$\nu$}}}                                  
\def\bphi{\mbox{\boldmath{$\phi$}}}
\def\bmu{\mbox{\boldmath{$\mu$}}}

\def\bDelta{\mbox{\boldmath{$\Delta$}}}
\def\bOmega{\mbox{\boldmath{$\Omega$}}}
\def\bPhi{\mbox{\boldmath{$\Phi$}}}
\def\bLambda{\mbox{\boldmath{$\Lambda$}}}
\def\bSigma{\mbox{\boldmath{$\Sigma$}}}
\def\bGamma{\mbox{\boldmath{$\Gamma$}}}
                                  
\newcommand{\myprob}[1]{\mathop{\mathbb{P}}_{#1}}

\newcommand{\myexp}[1]{\mathop{\mathbb{E}}_{#1}}

\newcommand{\mydelta}[1]{1_{#1}}

\newcommand{\myminimum}[1]{\mathop{\textrm{minimum}}_{#1}}
\newcommand{\mymaximum}[1]{\mathop{\textrm{maximum}}_{#1}}    
\newcommand{\mymin}[1]{\mathop{\textrm{minimize}}_{#1}}
\newcommand{\mymax}[1]{\mathop{\textrm{maximize}}_{#1}}
\newcommand{\mymins}[1]{\mathop{\textrm{min.}}_{#1}}
\newcommand{\mymaxs}[1]{\mathop{\textrm{max.}}_{#1}}  
\newcommand{\myargmin}[1]{\mathop{\textrm{argmin}}_{#1}} 
\newcommand{\myargmax}[1]{\mathop{\textrm{argmax}}_{#1}} 
\newcommand{\myst}{\textrm{s.t. }}

\newcommand{\denselist}{\itemsep -1pt}
\newcommand{\sparselist}{\itemsep 1pt}

\definecolor{pink}{rgb}{0.9,0.5,0.5}
\definecolor{purple}{rgb}{0.5, 0.4, 0.8}   
\definecolor{gray}{rgb}{0.3, 0.3, 0.3}
\definecolor{mygreen}{rgb}{0.2, 0.6, 0.2}

\newcommand{\cyan}[1]{\textcolor{cyan}{#1}}
\newcommand{\red}[1]{\textcolor{red}{#1}}  
\newcommand{\blue}[1]{\textcolor{blue}{#1}}
\newcommand{\magenta}[1]{\textcolor{magenta}{#1}}
\newcommand{\pink}[1]{\textcolor{pink}{#1}}
\newcommand{\green}[1]{\textcolor{green}{#1}} 
\newcommand{\gray}[1]{\textcolor{gray}{#1}}    
\newcommand{\mygreen}[1]{\textcolor{mygreen}{#1}}    
\newcommand{\purple}[1]{\textcolor{purple}{#1}}       

\definecolor{greena}{rgb}{0.4, 0.5, 0.1}
\newcommand{\greena}[1]{\textcolor{greena}{#1}}

\definecolor{bluea}{rgb}{0, 0.4, 0.6}
\newcommand{\bluea}[1]{\textcolor{bluea}{#1}}
\definecolor{reda}{rgb}{0.6, 0.2, 0.1}
\newcommand{\reda}[1]{\textcolor{reda}{#1}}

\def\changemargin#1#2{\list{}{\rightmargin#2\leftmargin#1}\item[]}
\let\endchangemargin=\endlist
                                               
\newcommand{\cm}[1]{}

\newcommand{\mhoai}[1]{{\color{blue}{[MH: #1]}}}

\newcommand{\mtodo}[1]{{\color{red}$\blacksquare$\textbf{[TODO: #1]}}}
\newcommand{\myheading}[1]{\vspace{0 ex}\noindent \textbf{#1}}
\newcommand{\htimesw}[2]{\mbox{$#1$$\times$$#2$}}


\newif\ifshowsolution
\showsolutiontrue

\ifshowsolution  
\newcommand{\Comment}[1]{\paragraph{\bf $\bigstar $ COMMENT:} {\sf #1} \bigskip}
\newcommand{\Solution}[2]{\paragraph{\bf $\bigstar $ SOLUTION:} {\sf #2} }
\newcommand{\Mistake}[2]{\paragraph{\bf $\blacksquare$ COMMON MISTAKE #1:} {\sf #2} \bigskip}
\else
\newcommand{\Solution}[2]{\vspace{#1}}
\fi

\newcommand{\truefalse}{
\begin{enumerate}
	\item True
	\item False
\end{enumerate}
}

\newcommand{\yesno}{
\begin{enumerate}
	\item Yes
	\item No
\end{enumerate}
}

\newcommand{\Sref}[1]{Sec.~\ref{#1}}
\newcommand{\Eref}[1]{Eq.~(\ref{#1})}
\newcommand{\Fref}[1]{Fig.~\ref{#1}}
\newcommand{\Tref}[1]{Table~\ref{#1}}

\definecolor{mygray}{gray}{.9}
\makeatletter
\newcommand{\thickhline}{%
    \noalign {\ifnum 0=`}\fi \hrule height 0.8pt
    \futurelet \reserved@a \@xhline
}
\newcommand{\sthickhline}{%
    \noalign {\ifnum 0=`}\fi \hrule height 1pt
    \futurelet \reserved@a \@xhline
}

\newlength{\DepthReference}
\settodepth{\DepthReference}{g}

\newlength{\HeightReference}
\settoheight{\HeightReference}{T}

\newlength{\Width}%

\newcommand{\MyColorBox}[2][red]%
{%
    \settowidth{\Width}{#2}%
    \colorbox{#1}%
    {%
        \raisebox{-\DepthReference}%
        {%
                \parbox[b][\HeightReference\DepthReference][c]{\Width}{\centering#2}%
        }%
    }%
}
\footnotetext[1]{Work done when interning at Meta}
\begin{abstract}

We introduce HOIGPT, a token-based generative method that unifies 3D hand-object interactions (HOI) perception and generation, offering the first comprehensive solution for captioning and generating high-quality 3D HOI sequences from a diverse range of conditional signals (\eg text, objects, partial sequences). At its core, HOIGPT utilizes a large language model to predict the bidrectional transformation between HOI sequences and natural language descriptions. Given text inputs, HOIGPT generates a sequence of hand and object meshes; given (partial) HOI sequences, HOIGPT generates text descriptions and completes the sequences. To facilitate HOI understanding with a large language model, this paper introduces two key innovations: (1) a novel physically grounded HOI tokenizer, the hand-object decomposed VQ-VAE, for discretizing HOI sequences, and (2) a motion-aware language model trained to process and generate both text and HOI tokens. Extensive experiments demonstrate that HOIGPT sets new state-of-the-art performance on both text generation (+2.01\% R Precision) and HOI generation (-2.56 FID) across multiple tasks and benchmarks.

\end{abstract}    
\section{Introduction}
\label{sec:intro}

\begin{figure}
\vspace{-0.2in} 
\centering
\includegraphics[width=1\linewidth]{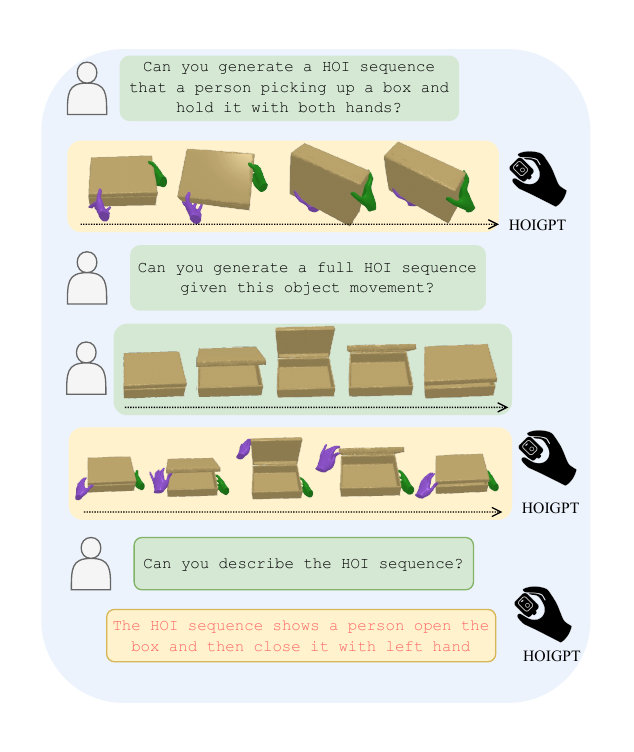}
\vspace{-0.45in} 
\caption{\textbf{HOIGPT} can interpret a variety of input prompts for diverse HOI-related tasks. We illustrate examples of text to HOI generation, HOI completion conditioned on object movement, and HOI captioning. HOIGPT generates or interprets hand-object interaction sequences in response to user queries, showcasing its capability to understand and produce contextually relevant HOI motions. The sequences represent time order from left to right.}
\label{fig:teaser}
\vspace{-0.25in}
\end{figure}

Hand-object interaction (HOI) refers to how humans use one or both hands to directly manipulate objects, including for tool manipulation, skilled activity, or exploration purposes.
These kinds of actions are among the primary ways that humans take action in the world.
Generating and understanding HOI in 3D is thus an important task pivotal to advancing machine capabilities in human-centric applications, such as human activity understanding, AR/VR, and robotics. 

A recent research trend has been the generation of HOI motion sequences from text~\cite{christen2024diffh2o, cha2024text2hoi}, primarily based on diffusion models~\cite{sohl2015deep}.
Such models are capable of generating HOI motion sequences with realistic appearances, but they have certain shortcomings that prevent them from being applied in broader HOI settings: 1) Limited flexibility in conditioning: Text prompts are often the only method of steering generation. This limits their ability to condition the motion as needed for motion completion, prediction, or blending; 2) Limited to short sequences: Diffusion models struggle to generate coherent, longer sequences of motion that includes multiple actions, partially due to challenges in leveraging prior knowledge and enforcing temporal consistency. Moreover, these works only focus on text-to-3D HOI generation and lack the capability to summarize actions, which restricts their applicability for other tasks.


In a seemingly unrelated field, large language models (LLMs) have made stunning recent advances~\cite{devlin-etal-2019-bert, t5, dubey2024llama, achiam2023gpt}. 
While not perfect, LLMs can now generate long, coherent text or even converse in multi-turn dialogue.
Interestingly, LLMs have also proven highly compatible with modalities beyond text, demonstrating increasing viability as universal interfaces and enabling transferr of the extensive knowledge distilled in pretrained LLMs to other modalities.
The latter is particularly attractive for HOI due to the difficulty of data collection (and subsequent small scale of existing data). Other characteristics of multimodal LLMs, such as bidirectionality, descriptive abilities, flexible conditioning, and long-sequence coherence, pose a potential solution to prior text-conditioned diffusion approaches to HOI.

Recognizing the potential of combining large language models (LLMs) with motion data, several recent studies have utilized LLMs to model human body motion~\cite{jiang2024motiongpt,zhang2024motiongpt, radosavovic2024humanoid, ribeiro2024motiongpt, wang2024motiongpt}. These approaches involve tokenizing motion data (e.g., using a VQVAE~\cite{van2017neural}) and fine-tuning the LLM to interpret these motion tokens as a novel ``language.''
A similar methodology can be applied to LLMs and HOI, but several challenges unique to HOI emerge.
Chief among these is that while human body motion effectively models a single entity, HOI involves the interaction of up to three. Modeling these interactions requires a next-level reasoning ability in physical world.
Furthermore, the set of possible objects in HOI is very diverse, each with varying interaction patterns and affordances, leading to virtually limitless possibilities; this variability in objects contrasts with human bodies and hands, which are often parameterizable with specialized and compact models~\cite{MANO, SMPL2015}.
The complex interplay between hands and objects and the objects' high diversity results in an enormous joint space of possible movements, which naive tokenization implementations can struggle to model.

In this paper, we propose \textbf{HOIGPT}, an autoregressive generative model designed for bidirectional transformations between 3D HOI motion and corresponding textual descriptions.
HOIGPT contains several key innovations.
Core to our work is a novel tokenizer specifically tailored for HOI that effectively factorizes HOI motion token space, reducing the required complexity and improving generality while still enforcing physical and semantical constraints to 3D hand-object interactions—a challenge not present in single-entity human body motion yet critical for generating plausible and coherent long HOI sequences. 
We also propose a pretraining approach for HOI tokens that enables the language model to capture the relationships between various HOI tokens more effectively. This enables several downstream HOI generation and understanding tasks, including HOI comprehension, generation, motion completion, and motion prediction—all using the same model with different prompts.

Our contributions are summarized as follows:
\begin{itemize}
    \item We introduce the first unified model for bidirectional transformation between HOI motion and text through autoregressive language models, enabling a single model to understand, generate, predict, and complete HOI motions.
    \item We present a HOI tokenizer, the first tokenizer designed to discretize HOI motions, addressing the lack of physical grounding in language tokens. Our approach incorporates novel geometric loss functions and a HOI-decomposed VQ-VAE architecture to enhance physical grounding.
    \item With these innovations, HOIGPT achieves state-of-the-art performance in both HOI generation and understanding, significantly outperforming all previous methods.
\end{itemize}


\section{Related Work}
\label{sec:formatting}

\myheading{Human motion generation with Multi-Modal Language Models}. 
Text to human motion generation~\cite{tevet2022motionclip, zhai2023language, guo2022generating, azadi2023make, zhang2024motiondiffuse,chen2023executing,dai2024motionlcm,wang2023fg,zhong2023attt2m,lou2023diversemotion} has become a well-studied task in recent years.
Among those methods, Diffusion Models~\cite{ho2020denoising} has been widely used for human motion synthesis \cite{cohan2024flexible, shafir2024human, tevet2023human}. However, those methods only generate human motion from text descriptions but do not generate text descriptions to demonstrate an understanding of hand motions.
Recently, the language model based motion generators are proposed to also achieve both generation and understanding of human motion.
\citet{zhang2023generating} proposed a general framework to generate human motion and text description based on VQ-VAE~\cite{van2017neural} and Generative Pre-trained Transformer (GPT)~\cite{brown2020language}. 
\citet{zhang2024motiongpt} propose to first tokenize human motion with a VQ-VAE~\cite{van2017neural} and then further finetune LLaMA-13B~\cite{touvron2023llama} with LoRA~\cite{hu2021lora}, utilizing both motion data and textual descriptions to enhance the generation process. 
\citet{jiang2024motiongpt} further proposed a unified framework for both motion synthesis and motion-to-text generation by full finetuning of a tiny language model~\cite{t5}.

\begin{figure*}[t]
\centering
\includegraphics[width=0.95\linewidth]{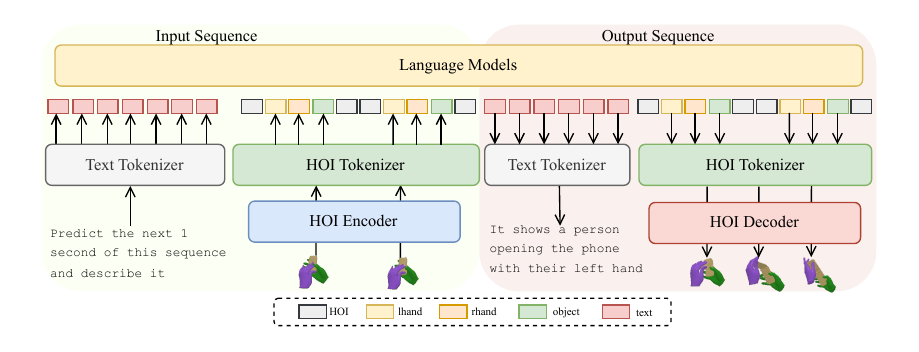} \hskip 0.1in
\vspace{-2em}
\caption{  
\textbf{Overview of the HOIGPT framework for bi-directional hand-object interaction (HOI) generation and understanding.} The input sequence (left) includes both text and HOI sequences, processed by the text tokenizer and HOI encoder, respectively. The HOI encoder uses an HOI Tokenizer to decompose HOI sequences into object, left hand, and right hand tokens. The language model takes both text and HOI tokens to generate the output sequence, which includes both text descriptions and generated HOI sequences. This design enables seamless integration of text and HOI data for tasks like motion prediction, description, and completion.
}  %
\label{fig:overview}
\vspace{-0.2in}
\end{figure*}

\myheading{Hand object interaction modeling}. 
Learning realistic hand-object interactions~\cite{liu2021semi,liu2022joint, jiang2021graspTTA, grauman2024egoexo4d} has been a longstanding challenge.
Recent advances have studied hand-object interactions by incorporating 3D representations of both hands and objects where the hand representations are MANO~\cite{MANO} parameters and objects are in 3d meshes. 
As capturing the intricate spatial relationships and physical constraints between hands and objects remains difficult, early pioneers are focusing on generating a realist grasping hand pose and predicting the affordance~\cite{jiang2021graspTTA}.
\citet{hasson2019learning} leveraged the hand-centric physical constraints for modeling the interaction between
hand-object to avoid penetration. Recently, many other works have focused on predicting hand object interactions~\cite{grauman2024egoexo4d} and~\citet{liu2022joint} proposed jointly predicting interaction hotspots and future trajectory and affordance.


\myheading{Text to HOI generation}. 
Early work~\cite{ghosh2023imos} focuses on generating a HOI sequence from an action label, but the resulting sequence is limited to a single action.
Recently HOI generation methods have adopted diffusion-based networks to generate HOI sequences from text descriptions. 
\citet{christen2024diffh2o} introduced a two-stage process generating HOI directly from text inputs, separating the task into grasping and interaction stages, with a transition refinement step bridging the two.
Text2HOI~\cite{cha2024text2hoi} also introduces a multi-step pipeline for generating HOI sequences from text descriptions, employing a diffusion-based approach conditioned on text prompts, affordance predictions, and object features to produce HOI sequences. The generated HOI is further optimized with a geometric loss to ensure more accurate and physically plausible results. 
While diffusion-based methods achieve state-of-the-art performance in HOI generation, they have certain limitations: (1) they rely on post-refinement stages to address physically implausible interactions, and (2) they are constrained by diffusion models~\cite{ho2020denoising}, which struggle to generate longer sequences. In contrast, our approach integrates geometric constraints directly into the training process, ensuring the generation of physically plausible HOI sequences without post-processing. Additionally, by leveraging large language models (LLMs), our method enables the generation of longer, more complex, and temporally consistent HOI sequences, surpassing the capabilities of existing methods.

\begin{figure*}[t]
\centering
\includegraphics[width=0.95\linewidth]{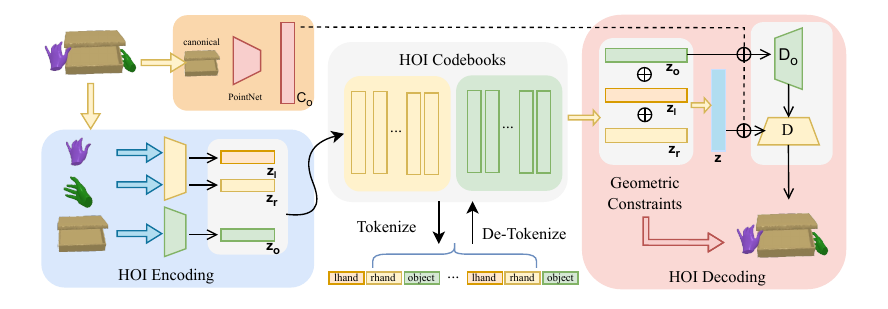} \hskip 0.1in
\vspace{-2em}
\caption{ \textbf{Overview of HOI-decomposed VQ-VAE.} Our framework processes hand and object features through dedicated hand and object encoders, which generate encoded representations. These representations are quantized using separate hand and object codebooks, resulting in corresponding codebook indices for each modality. The quantized indices are combined to form the HOI latent code, which is then decoded through object and hand decoders to reconstruct the HOI sequence. The reconstructed sequence captures realistic hand-object interactions that align closely with the input features. To further enhance physical plausibility, a geometric loss is applied, minimizing interpenetration between the hand and object and ensuring consistent, plausible contact dynamics.}  %
\label{fig:pipe}
\vspace{-3mm}
\end{figure*}

\section{Methodology}
\subsection{Overview}
We propose HOIGPT, a bidirectional text and 3D HOI motion generative model.
As with other recent multimodal LLM works~\cite{liu2023visual, moon2024anymal, jiang2024motiongpt}, we choose to unify both modalities in an LLM's input/output token space, letting us leverage the LLM's distilled knowledge and use it as a general purpose reasoning engine (Sec.~\ref{sec:llm}).
We map HOI motion sequences to the LLM's token domain through a novel HOI tokenizer, with codebooks that factorize hand motions from those of objects (Sec.~\ref{sec:hoi_token}).
We learn to index these HOI codebooks by learning a VQ-VAE that encode and decode HOI motion sequences into tokens from the codebooks and back again (Sec.~\ref{sec:vqvae}).
We impose several geometric regularizers during training of the VQ-VAE to ensure physical plausibility, and pre-train and instruction tune the LLM to learn to interpret the tokenized HOI motion (Sec.~\ref{sec:training}).

\myheading{Input/Output format}. We represent joint hand-object motion in 3D as a 6 degrees of freedom (6DoF) object pose and hand poses based on the MANO hand model~\cite{MANO}. Specifically, object poses are defined as $\mO=\left(\tau_o, \phi_o, \alpha_o\right)$ where $\alpha_o$ denotes the object articulation angle, $\tau_o$ represents the global 3D object location, and $\phi_o$ is the 6D object rotation~\cite{Zhou2018OnTC}. 
In addition, following previous work~\cite{cha2024text2hoi}, the geometric information (\eg point cloud) is provided as an input.
Similarly, hand poses are expressed as $\mathcal{H}=\left(\boldsymbol{\tau}, \boldsymbol{\phi}\right)$, where $\boldsymbol{\tau}$ is the global 3D hand location, and $\phi$ refers to the 6D hand rotations~\cite{Zhou2018OnTC}, derived from the MANO~\cite{MANO}  parameters, which can also output hand vertices. The final HOI motion sequences are represented as $X_s = \texttt{concat}(\mO, \mH_l, \mH_r)$. 
Finally, as text is the native input/output format for LLMs, we use the standard text tokenization process.

\subsection{HOI Language Model} \label{sec:llm}
We use a large language model (LLM) as HOIGPT's core processing unit, inheriting many of modern LLMs' language comprehension, instruction-following, and description generation capabilities.
Broadly speaking, the LLM in HOIGPT operates the same way as a standard language-only LLM: tokens in, tokens out.
More formally, given an input token sequence $\mT_s = \{ \mathbf{T}_s^i \}_{i=1}^N$ by tokenizing the HOI sequence $X_s$, the model autoregressively predicts the output sequence $\mT_t = \{ \mathbf{T}_t^i \}_{i=1}^L$, with the model learning to maximize the log-likelihood of the data distribution:
\begin{equation}
    \mathcal{L}_{lm} = - \sum_{i=0}^{L_t-1} \log p_\theta \left( \mathbf{T}_t^i \mid \mathbf{T}_t^{<i}, \mT_s \right).
\end{equation}
The key difference between HOIGPT and language-only LLMs is what these tokens represent.
While some tokens correspond to the typical text data, as is the case for the input prompt or output text (\eg a description), the rest of the tokens in HOIGPT's token sequences correspond to motion.
Training an HOI language model to interpret and generate such HOI motion tokens bestows HOIGPT's versatility to correspondingly interpret and generate HOI motion.
Below, we outline the process of achieving a mapping from HOI motion sequences into tokens.

\subsection{Hand and Object Motion Codebooks} \label{sec:hoi_token}
For HOIGPT's language model to interpret 3D HOI motion sequences, they must first be represented as a form of ``language,'' which the LLM can later be finetuned to learn to interpret.
We define the ``vocabulary'' of motion in the form of a fixed-size codebook containing a bidirectional mapping of embeddings to discrete token IDs, similar to that of text.
A similar strategy has previously been employed by several works studying full human body pose sequences~\cite{wang2024motiongpt, jiang2024motiongpt, zhang2024motiongpt, ribeiro2024motiongpt}, compressing and quantizing the space of human actions into a dictionary.
However, a key difference is that there is ultimately only one entity to model for human body pose motion sequences.

In contrast, HOI involves up to three entities: two hands and an object.
This requires an improved design: A naive approach of compressing the joint motion of three semi-independent units into a single codebook scales combinatorially and is thus unnecessarily wasteful.
For example, one-handed actions leave the idle hand as a free variable; it's both inefficient and unnecessary to have a codebook entry for every possible motion or position of the non-acting hand.
Similarly, HOI encapsulates a wide variety of potential objects, yet certain canonical motions (\eg lifting, opening) are highly transferable across objects. 
Given these considerations, we have chosen to factorize the HOI codebook as a hand codebook and an object codebook. 
We choose a single-hand codebook to represent both hands, as almost all motions by one hand can be mirrored by the other, for which there's little reason to re-learn a separate codebook entry; we instead separately encode handedness (left or right).

Given these codebooks and a set of hand and object motion features (see Sec.~\ref{sec:vqvae}), tokenization of an HOI motion sequence is straightforward. 
We do a nearest-neighbor look-up of features for each hand and object in the corresponding codebook, yielding a token index. 
We sandwich the HOI motion tokens with a special $\langle\texttt{HOI}\rangle$ token, yielding the following sequence to be passed to the language model:
\begin{align*}
    \small{
    \langle \texttt{HOI} \rangle \langle \mathrm{H}_1^{(L)} \rangle \langle \mathrm{H}_1^{(R)} \rangle \langle \mathrm{O}_1 \rangle ... \langle \mathrm{H}_T^{(L)} \rangle \langle \mathrm{H}_T^{(R)} \rangle \langle \mathrm{O}_T \rangle \langle \texttt{HOI} \rangle
    }
\end{align*}
where $\langle\mathrm{H}_t^{(L/R)}\rangle$ represent the $t^{\mathrm{th}}$ time window hand token for the left (L) or right (R) hand and $\langle\mathrm{O}_t\rangle$ the object's. 
For HOI motion generation, the process is reversed: the special $\langle\texttt{HOI}\rangle$ tokens are stripped, and then the predicted HOI motion token indices are used to look up a latent embedding for the left/right hand and object in the appropriate codebook.

\subsection{HOI-decomposed VQ-VAE} \label{sec:vqvae}
The final architectural component of HOIGPT is a VQ-VAE~\cite{van2017neural} which encodes from and decodes into 3D HOI motion sequences, and whose posterior $q(z|x)$ is represented by the aforementioned hand and object motion codebooks. Different from previous work~\cite{cha2024text2hoi} that jointly models the hand and object motion sequences, we quantize the HOI sequence by decomposing hands and object. 

\myheading{HOI Encoder.} HOIGPT's encoder consists of three feature extractors: hand motion, object motion, and object point cloud. 
First, the object point cloud sampled from canonical object mesh vertices is encoded using PointNet~\cite{qi2017pointnet}, resulting in features $\mathbf{c}_o$ representing the shape of object, which helps narrow the types of possible HOI motion.
We thus use it as conditioning for the two hand and object feature extractors, which independently project a windowed motion sequence for an entity to a corresponding latent vector: hands features $\mathbf{z}_l$ and $\mathbf{z}_r$, and object features $\mathbf{z}_o$.
These features $\{\mathbf{z}_l, \mathbf{z}_r, \mathbf{z}_o\}$ are then used to index into the appropriate HOI motion codebook (as described in Sec.~\ref{sec:hoi_token}) to generate HOI tokens corresponding to the input HOI motion.

\myheading{HOI Decoder.} The HOI decoder operates on four latent feature sets: reconstructed hand features $\mathbf{\hat{z}}_l$ and $\mathbf{\hat{z}}_r$, reconstructed object features $\mathbf{\hat{z}}_o$, and point cloud features $\mathbf{c}_o$. 
The point cloud features $\mathbf{c}_o$ are retained from the HOI Encoder's canonical point cloud feature extractor, as described earlier. 
Meanwhile, the reconstructed latent features $\{\mathbf{\hat{z}}_l, \mathbf{\hat{z}}_r, \mathbf{\hat{z}}_o\}$ are either directly obtained from the encoder during reconstruction, or generated by the HOI LLM in response to a query.
To decode these feature, we first add the object motion and point cloud features $\mathbf{\hat{z}}_o + \mathbf{c}_o$ and predict the object motion alone with object decoder $\mD_o$.
We then combine all three features $\mathbf{z}= \mathbf{\hat{z}}_o + \mathbf{\hat{z}}_l + \mathbf{\hat{z}}_r$, and decode the hand-object interaction with the hand decoder $\mD$, conditioned on the predicted object motion. 
Conditioning the hand motion on the predicted object motion ensures that the generated hand movements respect the object's spatial and geometric properties. 
The final output HOI sequence $\hat{X}_s$ is defined as:
\begin{equation}
    \hat{X}_s = \mD( (\mathbf{\hat{z}}  + \mathbf{c}_o)| \mD_o(\mathbf{\hat{z}}_o + \mathbf{c}_o))
\end{equation}

\subsection{Training} \label{sec:training}
\vspace{-1mm}
\myheading{HOI-decomposed VQ-VAE and tokenizer}.
Following~\cite{van2017neural, zeghidour2021soundstream}, we incorporate reconstruction and latent embedding losses at each quantization stage. To ensure accurate reconstruction of both hand and object motions, an embedding loss is computed for each latent representation. Different from RVQ~\cite{zeghidour2021soundstream}, our method utilizes a dual HOI encoder, decomposing hands and object, enabling supervised learning of the residual.
The following loss is used as supervision:
\begin{equation}
\mathcal{L}_{\text{tok}} = |X - \hat{X}|_1 + \alpha \sum_i^I|\mathbf{z}_i - \operatorname{sg}[\hat{\mathbf{z}}_i]|_2^2, ~I \in \{o,l,r\}
\end{equation}
where sg[$\cdot$] denotes the stop-gradient operation, and 
$\alpha$ is the weighting factor for the embedding constraint. This embedding loss is essential for minimizing reconstruction errors and maintaining the fidelity of each decomposed part for hand, object and their interactions. By decomposing hands and object, our proposed method captures the nuances of hand-object dynamics, including contact points and relative motions, making it highly effective for detailed HOI generation  tasks.
To further enhance the robustness of HOI reconstruction, we follow the masking strategy from MAE~\cite{he2022masked}, where the hand or object latents are randomly masked. The masked latent is set to zero, diminishing its contribution. This design encourages each quantized latent for the hands or object to also learn correlations with the other, fostering stronger interactions between the components.

\myheading{Geometry-aware tokenization}. 
We introduce a novel geometric-aware loss to train HOI tokenization. Unlike other modalities that lack interactions between multiple entities, HOI requires enforcing realistic interactions between hands and objects. This requires minimizing physically implausible penetrations while maintaining supportive contact between hand and object surfaces. The geometric-aware loss is a combination of the following three geometric losses.
\\
\textbf{Penetration Loss} $(\mL_{pen})$. Following~\cite{jiang2021hand}, the penetration loss is employed to penalize predicted hand poses $\hat{\mH}$ where hand vertices penetrate the object’s surface. Specifically, the MANO model~\cite{MANO} converts the predicted hand pose $\hat{\mH}$ into hand vertices $\mathcal{P}$. For each hand vertex $P \in \mathcal{P}_{\text{in}}^o$ that penetrates the object, we find the closest object vertex \( {V}^o_i \) and minimize the squared distance \( \|P - {V}^o_i\|_2^2 \). The penetration loss is then averaged over all penetrating hand vertices, discouraging unrealistic penetrations:
\begin{equation}
    \mL_{\text{pen}} = \frac{1}{|\mathcal{P}_{\text{in}}^o|} \sum_{p \in P_{\text{in}}^o} \min_i \| p - \hat{V}_i \|_2^2
\end{equation}
\\
\textbf{Contact grasping loss} $(\mL_C)$. The contact approach loss ensures that the hand joints remain within a specified distance \( \phi \) from the nearest object surface, maintaining realistic motion trajectories before contacting. Specifically, \( D(J_i^o) \) calculates the distance between hand joint and the closest object surface, enforcing the constraint \( D(J_i^o) \leq \phi \):
\begin{equation}
    \mL_C = \sum_{i} D(\hat{J}_i^o), \quad \forall D(\hat{J}_i^o) \leq \phi
\end{equation}
where \( D(\hat{J}_i^o) = \min_j \| V_j^p - \hat{J}_i^o \|_2^2 \), with \( V_j^p \) representing the object surfaces and \( J_i^o \) representing the hand joints. This loss function facilitates smooth 
transitions between the hand's approach and its initial contact with the object.
\\
\textbf{Contact region loss} $(\mL_R)$. 
The contact region loss promotes physically plausible contact by aligning ground truth hand vertices \( P_i^o \) at contact locations with corresponding reconstructed vertices \( \hat{P}_i^o \) and ensures that hand vertices remain within a specified distance \( \tau \) from the nearest object surface, maintaining realistic interaction without gaps. This loss minimizes the distance between the ground-truth hand vertices within a threshold distance \( \tau \) of the object surface and their reconstructed counterparts and encourage the contact between hand and object, ensuring realistic hand-object contact points by optimizing the predicted affordance region: 
\begin{equation}
    \mL_R = \sum_{i} \| \Phi(P_i^o) - \Phi(\hat{P}_i^o) \|_2 + D(\hat{P}_i^o),  \quad \forall D(P_i^o) < \tau    
\end{equation}
where $\Phi(\cdot)$ is a binary function that indicates whether $\D(\cdot)$ is less than $\tau$, and \( D({P}_i^o) \) is the distance between hand vertices and the nearest object surface. The overall geometric loss is formulated as a weighted sum of those three losses:
    \begin{equation}
    \mL_{geo} = \lambda\mL_{pen} + \beta\mL_C + \gamma\mL_R    
\end{equation}
The geometric constraints improve the quality of hand-object interactions by penalizing unnatural penetrations, enforcing proximity between the hand and object before contacting, and ensuring contact at physically plausible regions.


\myheading{HOI tokenization loss.} The final training objective function is formulated as:
    \begin{equation}
    \mL = \mL_{geo} + \mL_{tok}
\end{equation}

\myheading{Language model}. We train the HOI language model in two stages: In the \textbf{Motion-language Pre-training Stage}, we finetune the T5 model~\cite{colin2020exploring, t5} on a combination of language and motion data through both unsupervised and supervised methods. By randomly masking tokens in the input, we train the model to predict these spans, enabling it to learn the relationship between motion and language from paired datasets. Following this, in the \textbf{Instruction Tuning Stage}, we construct a multi-task text-motion dataset with instruction-based prompts, covering tasks such as motion generation and captioning. This instruction tuning enhances the model's ability to generalize to unseen prompts and tasks, thereby improving its performance across diverse motion-language applications.

\section{Experiments} %
\label{sec:exp}

\begin{table*}[t]
    \centering
    \resizebox{0.82\linewidth}{!}{
    \begin{tabular}{lcccccc}
        
    Methods & FID $\downarrow$& Diversity$\rightarrow$ & RPrecision (Top 3)$\uparrow$ & MMDist$\downarrow$ & MModality$\uparrow$ & IV$\downarrow$\\
    \sthickhline 
    Real & 0.01 & 17.03 & 86.99 & 2.05 & - & -\\
    \hline
    T2MGPT~\cite{zhang2023generating} &8.69 {\small $\pm$ 1.06} & 12.69 {\small $\pm$ 1.33} & 68.99 {\small $\pm$ 0.01} & 8.97 {\small $\pm$ 0.13} & 4.71 {\small $\pm$ 0.18} & 14.3\\
    MotionGPT~\cite{jiang2024motiongpt} &8.71 {\small $\pm$ 0.46}& 11.25 {\small $\pm$ 1.81} & 63.67 {\small $\pm$ 0.02} & 6.10 {\small $\pm$ 0.20} & 4.17 {\small $\pm$ 0.18} & 16.3\\
    TM2T~\cite{guo2022tm2t} &9.51 {\small $\pm$ 0.46}& 10.72 {\small $\pm$ 2.02} & 64.80 {\small $\pm$ 0.09} & 6.30 {\small $\pm$ 0.09} & 4.40 {\small $\pm$ 0.17} & 15.3\\
    Text2HOI~\cite{cha2024text2hoi} & 5.85 {\small $\pm$ 0.46} & 15.43 {\small $\pm$ 1.69} & 72.34 {\small $\pm$ 0.02} & 5.24 {\small $\pm$ 0.22} & 1.66 {\small $\pm$ 0.21} & 12.3 \\
    
    \rowcolor[gray]{0.9}HOIGPT & \textbf{3.29} {\small $\pm$ 0.54} & \textbf{15.60} {\small $\pm$ 2.18} & \textbf{74.80} {\small $\pm$ 0.02} & \textbf{4.71} {\small $\pm$ 0.21} & \textbf{4.75} {\small $\pm$ 0.22} & \textbf{12.1}\\
    
    \end{tabular}
    }
    \caption{\textbf{Comparison with the state-of-the-art on HOI generation}. The arrows ($\rightarrow$) indicate that closer to real is desirable. The best performance are highlighted in \textbf{bold}.}
    \label{tab:comparison_text_to_hoi}
\end{table*}

\subsection{Implementation Details}
We use a codebook size of 512 for both the hand and object codebooks. We utilize the 220M parameter Flan-T5-Base~\cite{t5} as our language model.
The HOI tokenizer is trained for 2000 epochs, and the language model is fine-tuned for 100 epochs using a learning rate of $2e^{-4}$. Experimental results are calculated with a 95\% confidence interval from 20 repeated runs, ensuring statistical significance.  All experiments are conducted on 32 NVIDIA A100 GPUs.

\subsection{Dataset}
For our experiments, we combine two hand-object interaction datasets: ARCTIC~\cite{fan2023arctic} and GRAB~\cite{taheri2020grab}. ARCTIC includes 11 articulated objects, with various hand-object interactions. Text descriptions for this dataset are manually annotated by~\cite{cha2024text2hoi}.
GRAB focuses on general human-object interactions. For our experiments, we use the subset of GRAB containing only hand-object interactions, which contains 58 objects and diverse associated interactions. We generate the text descriptions based on the hand intent action labels and object name. Unlike previous works \cite{cha2024text2hoi}, we do not divide a long sequence with multiple actions into several single-action sequences. We automatically annotate the hand contact information (\eg right hand, left hand, or both) in the text descriptions by computing the distance between hand vertices and the object surface, applying a threshold to determine if contact occurs.
We reserve 500 unseen HOI sequences for testing, using the remaining 5.6k for training.


\begin{table*}[h!]
\vspace{-0.15in}
    \setlength{\tabcolsep}{13pt}
    \centering
    \resizebox{0.82\linewidth}{!}{
    \begin{tabular}{lcccc|ccc}
        \multirow{2}{*}{Methods} & \multicolumn{4}{c|}{HOI Prediction} & \multicolumn{3}{c}{HOI Interpolation} \\ 
        & FID $\downarrow$ & Diversity $\uparrow$ &

        ADE $\downarrow$ & FDE $\downarrow$ & FID $\downarrow$ & Diversity $\uparrow$ & ADE $\downarrow$ \\ 
        \thickhline
         Real &0.01&15.33& -  & - & 0.01 & 14.63 &- \\ 
         \hline
        MotionGPT~\cite{jiang2024motiongpt} &8.23&9.34&5.86&8.43 & 6.16 & 10.91 & 5.11 \\
        \rowcolor[gray]{0.9}HOIGPT & 4.08&12.59&4.79&6.84&  2.91 & 11.31 & 4.06 \\
    \end{tabular}
    }
    \vspace{-0.1in}
    \caption{\bf Comparison on HOI prediction and HOI interpolation. }
    \label{tab:compl}
    \vspace{-0.2in}
\end{table*}

\begin{table}[ht]
    \centering
    \resizebox{0.9\columnwidth}{!}{
    \begin{tabular}{lccc}
        
     & FID $\downarrow$& Diversity$\uparrow$  & IV$\downarrow$\\
    \thickhline 
    VQ-VAE tokenizer & 0.87 & 10.56 & 14.23\\
    w/o HOI-decomposed VQ & 0.53& 12.94 & 8.82\\
    w/o penetration loss & 0.64 & 11.52 & 12.32\\
    w/o contact loss &  0.67& 11.21 & 9.21\\
    \rowcolor[gray]{0.9}HOI Tokenizer (Ours) & 0.43&  13.62 & 8.60\\
    
    \hline
    \end{tabular}
    }
    \vspace{-0.1in}
    \caption{{\bf Ablation on tokenizer design and geometric constraints.}}
    \label{tab:abl}
    \vspace{-0.1in}
\end{table}

\subsection{Evaluation metrics}
\myheading{Text to HOI generation}.
To assess the capability of our model to generate long HOI sequences involving multiple actions, we adopt the evaluation protocol from~\cite{tevet2023human, guo2022generating}. The evaluation is conducted using the following metrics:
\begin{itemize}[noitemsep]
    \item \textbf{Fréchet Inception Distance (FID)} is our primary metric for HOI generation quality, measuring the distance between feature distributions of generated and real motions based on a feature extractor from~\cite{heusel2017gans}. 
    \item \textbf{Diversity} is used to measure the diversity of our generated motions, which calculates the variance within features extracted from the motions.
    \item \textbf{MultiModality (MModality)} measures the diversity of generated motions for a single text description, reflecting ability to produce varied outputs for the same prompt.
    \item \textbf{R Precision (motion-retrieval precision)} measures the matching accuracy between texts and motions using Top 3 retrieval accuracy.
    \item \textbf{Multi-modal Distance (MMDist)} evaluates the alignment between generated motions and corresponding texts by measuring the distance between them in a shared feature space, with lower values indicating closer alignment between the two modalities. 
    \item Besides standard metrics, we also follow \cite{hasson2019learning} to report the \textbf{interpenetration volume (IV)} by measuring the number of hand vertices that penetrate the object surfaces, as well as the maximum depth of interpenetration. 
\end{itemize}

\begin{table}[ht]
\centering
\scalebox{0.95}{
\begin{tabular}{lcccc}

\multirow{2}{*}{Methods} & \multicolumn{3}{c}{R Precision↑} & \multirow{2}{*}{MM Dist$\downarrow$}  \\
 & Top 1 & Top 2 & Top 3 &   \\
\thickhline 
Real  & {60.93} & {73.74} & {81.25} & {2.48}  \\
\hline
TM2T & 21.87 & 42.18 & 45.67 & 9.34\\
MotionGPT &24.68 &40.63& 46.88&8.83 \\
\rowcolor[gray]{0.9}HOIGPT & 26.69 & 42.19 & 48.43 &7.55  \\

\end{tabular}}
\caption{\bf Comparison of HOI to text generation approaches.}
\vspace{-0.1in}
\label{tab:m2t}
\end{table}

\begin{table}[ht]
    \centering
    \resizebox{0.95\columnwidth}{!}{
    \begin{tabular}{lcccc}
        
    HOIGPT & FID $\downarrow$& Diversity$\uparrow$  & RPrecision $\uparrow$ & IV $\downarrow$\\
    \thickhline 
    w/  obj-cond & 2.14 & 11.71 & 82.71 & 11.2\\
    w/o obj-cond & 3.29&  15.60 & 74.80 & 12.1\\
    
    \end{tabular}
    }
    \vspace{-0.1in}
    \caption{{\bf Conditioning HOIGPT on object motion.} HOIGPT can generate high fidelity HOI motion conditioned on object motion.}
    \label{tab:posec}
    \vspace{-0.2in}
\end{table}


\myheading{HOI to text generation.} To evaluate accuracy of HOI to text generation, we employ R Precision and MMDist as our primary metrics. 

\myheading{Motion prediction and interpolation.} For motion prediction and interpolation tasks, we use metrics commonly applied in motion prediction studies~\cite{yuan2020dlow, zhang2021we, ma2022multi}, including Average Displacement Error (ADE) and Final Displacement Error (FDE), to assess the accuracy of the predicted motion.

\subsection{Main results}
\myheading{Text to HOI generation}.
The text-to-HOI task involves generating hand-object interaction sequences from a text input. We evaluate our proposed HOIGPT model as the pre-trained model for the text-to-HOI task. For fair comparison, we finetune baseline methods on our combined HOI dataset. 
In Table~\ref{tab:comparison_text_to_hoi}, we compare HOIGPT's performance with state-of-the-art HOI generation method Text2HOI~\cite{cha2024text2hoi} and other language model based human motion synthesis baselines: T2MGPT~\cite{zhang2023generating}, MotionGPT~\cite{jiang2024motiongpt}, TM2M~\cite{guo2022tm2t}. HOIGPT significantly outperforms all other methods on all metrics.

\myheading{HOI to text generation}.
The HOI-to-text task involves generating a text description given a novel sequence of hand-object interactions. We benchmark HOIGPT against recent methods on the same HOI datasets, using established metrics to measure the accuracy and relevance of generated descriptions. The reported results from previous works use pre-processed ground truth descriptions, standardizing grammatical tense and plural forms to ensure consistency. In Table~\ref{tab:m2t}, we present a comparison of our method with  state-of-the-art approaches. These results indicate that HOIGPT consistently outperforms existing methods, generating more accurate descriptions of given hand-object interactions.

\myheading{HOI prediction and interpolation}. Following previous work~\cite{jiang2024motiongpt}, we evaluate HOIGPT on two additional downstream tasks,  HOI prediction and HOI interpolation. For the HOI prediction task, the model is conditioned on the first 20\% of HOI sequence, and the remaining 80\% is predicted.
In the HOI interpolation task, 50\% of the sequence is randomly masked, and the model is tasked with completing the missing segments. To evaluate performance, we employ FID, ADE, and FDE metrics and compare our method with \cite{jiang2024motiongpt} in \Tref{tab:compl}. The results demonstrate that HOIGPT achieves significantly improvements across all metrics.


\begin{figure*}[!t]
\vspace{-0.2in} 
\centering
\includegraphics[width=0.98\linewidth]{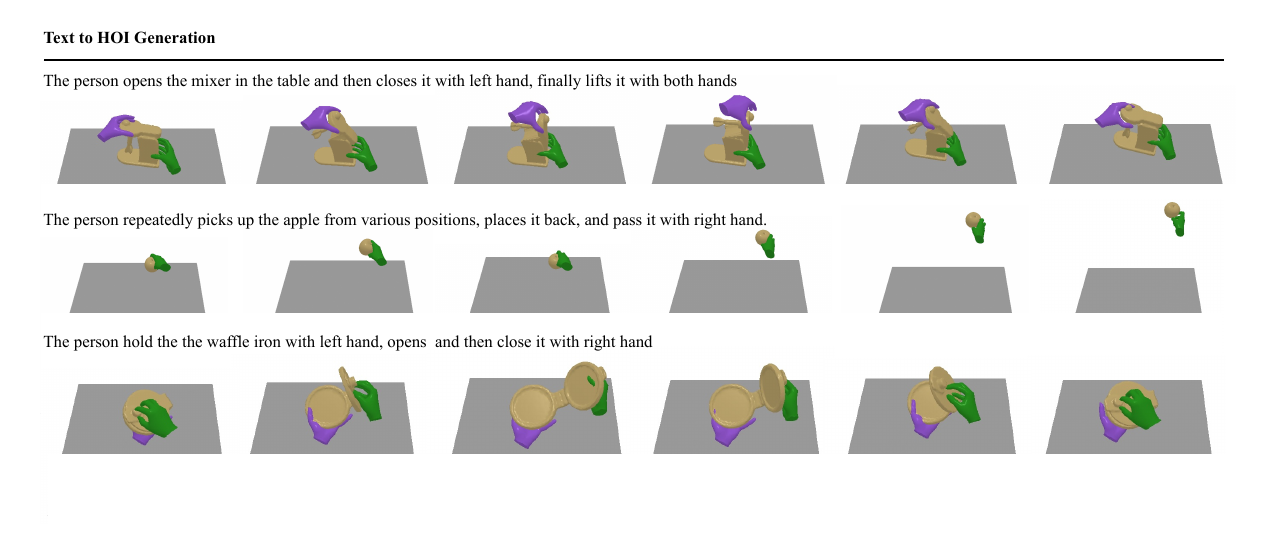}
\vspace{-0.6in} 
\caption{\textbf{Text to HOI generation examples.} HOIGPT generates long HOI sequences with multiple complex actions with only text input.}
\label{fig:quli}
\end{figure*}

\begin{figure*}[!t]
\vspace{-0.2in} 
\centering
\includegraphics[width=0.96\linewidth]{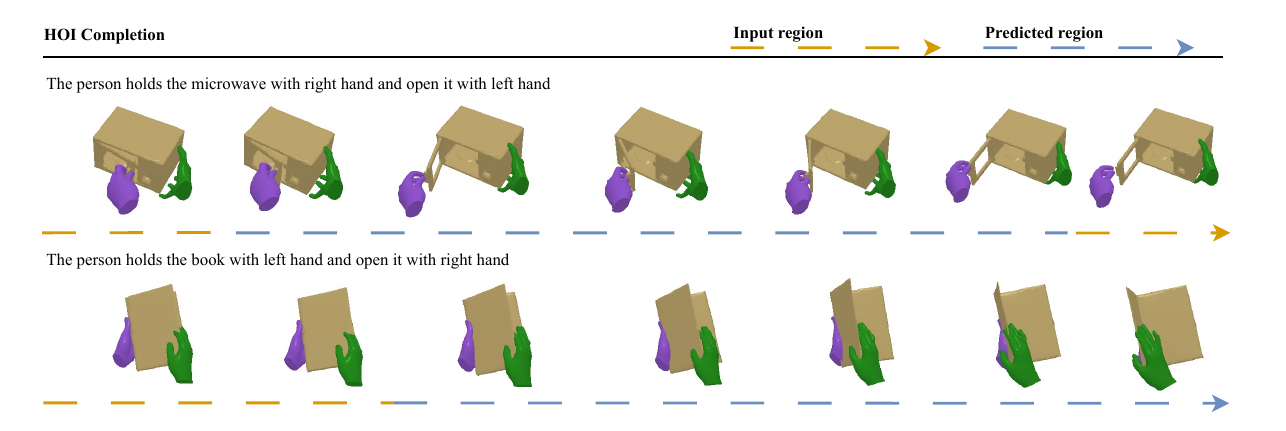}
\vspace{-0.2in} 
\caption{\textbf{Qualitative results of HOIGPT for HOI completion.} HOIGPT is designed for multiple tasks including HOI interpolation (top) and HOI prediciton (bottom), the \textcolor{orange}{orange} line indicts the input HOI sequence.}
\label{fig:quli2}
\vspace{-0.25in}
\end{figure*}
\subsection{Ablation Studies}

\myheading{Effectiveness of HOI-decomposed HOI Tokenization.}
We compared the HOI Tokenizer with a naive VQ-VAE model~\cite{van2017neural}, in which the hand and object components are separately quantized using two distinct VQ-VAE models. To further evaluate the impact of the HOI-decomposed VQ-VAE, we conducted an ablation study by removing this component. In the ablation setup, HOI sequences were processed directly without the decomposition and hierarchical tokenization of hand-object interactions, bypassing the structured quantization approach employed in our method.
As shown in \Tref{tab:abl}, the naive VQ-VAE model yielded the weakest performance, underscoring the challenges of discretizing HOI motion. In contrast, incorporating HOI-decomposed component improved performance by 23\%, demonstrating its effectiveness on capturing the complexity of hand-object interactions (\Tref{tab:abl}). 
Note that we reported the results for reconstructed HOI sequences without training the language model. We then used the best-performing model for subsequent language model training.


\myheading{Effectiveness of geometric losses.}
To evaluate the impact of the geometric loss on model performance, we conducted an ablation study by removing each geometric loss (contact-based loss and penetration loss) in HOIGPT training. The geometric loss enforces realistic hand-object interactions by minimizing interpenetration and promoting plausible contact.
As shown in Table~\ref{tab:abl}, removing the geometric loss results in increased interpenetration and less natural hand-object interactions, reducing the quality of generated HOI sequences and decreasing the accuracy of corresponding text descriptions. In contrast, with the geometric loss, the model produces high-quality, physically plausible HOI sequences with realistic contact, significantly improving performance in both text-to-HOI and HOI-to-text tasks.

\myheading{Object motion conditioned HOI generation.}
Beyond bidirectional text-to-HOI and HOI-to-text capabilities, our model HOI-decomposed tokenization also supports HOI generation conditioned on object motion. 
During training, we replace hand tokens with a mask token $\langle$\texttt{MASK}$\rangle$ and train the LLM to predict the masked tokens. At inference time, given HOI tokens with masked hand tokens as input, HOIGPT predicts the hand motion tokens, which are subsequently decoded into HOI sequences. 
By conditioning on object motion, our model learns to generate realistic and contextually appropriate hand movements that align with the object's behavior. As shown in \Tref{tab:posec}, this approach results in more plausible and synchronized HOI sequences, demonstrating the effectiveness of object-motion conditioning in enhancing the realism of hand-object interactions.

\subsection{Qualitative Results}
We also provide qualitative results in \Fref{fig:quli}, \Fref{fig:quli2} to highlight HOIGPT's ability to generate natural and contextually accurate hand-object interactions across various tasks. 
\section{Conclusion}
We have introduced HOIGPT, a novel token-based generative model designed to bridge the gap between 3D hand-object interactions (HOI) and natural language. Our approach leverages a unique HOI-decomposed VQ-VAE for HOI tokenization and a language model, enabling bidirectional transformation between HOI sequences and text descriptions. By incorporating HOI tokens within the VQ-VAE, our model captures complex interaction patterns and motion details, enhancing the richness and accuracy of HOI representations. Extensive experiments demonstrate that HOIGPT achieves state-of-the-art performance across multiple benchmarks, outperforming existing methods in multiple tasks. These results highlight the potential of our approach to improve understanding and generation of hand-object interactions, offering valuable applications in areas such as robotics, virtual reality, and human-computer interaction. Future works will explore scaling the model to more diverse datasets.
\clearpage
\newpage
{
    \small
    \bibliographystyle{ieeenat_fullname}
    \bibliography{main}
}



\end{document}